\documentclass{article}

\usepackage{arxiv}
\usepackage{hyperref}       
\usepackage{booktabs}       
\usepackage{amsfonts}       
\usepackage{amsmath}
\usepackage{graphicx}
\usepackage{doi}
\usepackage[ruled,vlined,linesnumbered]{algorithm2e}
\usepackage{tikz}
\usepackage{multirow}
\usepackage{longtable}
\usepackage[tableposition=below]{caption} 
\usepackage{caption}
\usepackage{spverbatim}
\usepackage{fancyvrb}

\DefineVerbatimEnvironment{VerbatimUnderline}{Verbatim}{commandchars=\\\{\}}

\usetikzlibrary{positioning}

\title{A General Retrieval-Augmented Generation Framework for Multimodal Case-Based Reasoning Applications}

\date{}

\newif\ifuniqueAffiliation
\uniqueAffiliationtrue
\author{
    Ofir Marom \\
    Independent Researcher \\
    \texttt{ofiremarom@gmail.com}
}

\renewcommand{\shorttitle}{\textit{arXiv} Template}

\begin{document}
\maketitle

\begin{abstract}
Case-based reasoning (CBR) is an experience-based approach to problem solving, where a repository of solved cases is adapted to solve new cases. Recent research shows that Large Language Models (LLMs) with Retrieval-Augmented Generation (RAG) can support the Retrieve and Reuse stages of the CBR pipeline by retrieving similar cases and using them as additional context to an LLM query. Most studies have focused on text-only applications, however, in many real-world problems the components of a case are multimodal. In this paper we present MCBR-RAG, a general RAG framework for multimodal CBR applications. The MCBR-RAG framework converts non-text case components into text-based representations, allowing it to: 1) learn application-specific latent representations that can be indexed for retrieval, and 2) enrich the query provided to the LLM by incorporating all case components for better context. We demonstrate MCBR-RAG's effectiveness through experiments conducted on a simplified Math-24 application and a more complex Backgammon application. Our empirical results show that MCBR-RAG improves generation quality compared to a baseline LLM with no contextual information provided.
 \end{abstract}

\section{Introduction}

Case-based reasoning (CBR) is an experience-based approach to problem solving, utilizing a repository of previously solved cases to address new problems through a four-phase pipeline: Retrieve, Reuse, Revise, and Retain \cite{aamodt1994, gilboa1995}. A solved case consists of a \textit{problem} and a \textit{solution}, while a new case includes only a problem. The first two phases aim to retrieve similar cases from the repository and adapt their solutions for the new problem. The Revise phase then ensures the correctness of the new solution, while the Retain phase expands the repository by incorporating newly solved cases.

Recent advancements in Large Language Models (LLMs), particularly those enhanced with Retrieval-Augmented Generation (RAG), demonstrate significant promise in supporting the Retrieve and Reuse stages of the CBR pipeline \cite{wiratunga2024, yang2024, guo2024, sourati2023}. While these studies employ varied methodologies tailored for specific applications, they fundamentally share the same core principles.

The Retrieve stage utilizes RAG retrieval, indexing solved case problems as neural embeddings. When a new case is presented, its problem is also converted to an embedding, enabling a similarity measure to identify related cases. The Reuse phase then employs RAG generation, crafting a query for the LLM to produce a solution for the new case while integrating knowledge from retrieved cases as additional context.

While promising results have emerged from this research, the focus has primarily been on text-only applications such as legal Q\&A, medical diagnostics, code generation, and logical fallacy detection \cite{wiratunga2024, yang2024, guo2024, sourati2023}. In such applications, case problems can be effectively converted to neural embeddings using pretrained transformer-based language models, such as BERT or GPT \cite{devlin2019, radford2019}. However, many CBR applications naturally encompass data other than text such as image, audio, and video formats \cite{schmidt2001, zeleznikow1993, portinale2004}.

In this paper we introduce MCBR-RAG, a general RAG framework designed for multimodal CBR applications. Previous research has explored multimodal RAG for various modalities, including image, audio, and video \cite{yasunaga2023, li2023, ghosh2023, xu2024}. These models typically encode multimodal documents into a latent space for retrieval, and subsequently transform them into text-based representations to provide context for LLM queries. Adopting this methodology, our framework accommodates cases with non-text modalities, requiring additional processing steps within the CBR pipeline.

Although models like BLIP-2 \cite{li2023} facilitate the conversion of multimodal documents to text, CBR applications often necessitate the development of application-specific models. In this paper, we present two multimodal CBR applications that exemplify this need: a simplified Math-24 domain which integrates text and image components to illustrate our framework's operation in a controlled setting, and a more complex Backgammon application that demonstrates the framework’s efficacy in addressing real-world tasks involving similar textual and visual interactions.

In summary, the contributions of this paper are twofold: first, we formalize the MCBR-RAG framework in an application-agnostic manner that is suitable for various multimodal CBR scenarios. Secondly, we illustrate the practical implementation of this framework by training the necessary text generation and latent representation models for our Math-24 and Backgammon applications.

The remainder of this paper is organized as follows: in Section \ref{background}, we provide additional background on CBR and RAG. In Section \ref{methodology}, we formalize our proposed MCBR-RAG framework. In Section \ref{applications}, we detail our Math-24 and Backgammon CBR applications. In Section \ref{experiments}, we run experiments showing that our framework improves generation quality over a baseline LLM that uses no additional context. We conclude with our final remarks in Section \ref{final-remarks}.

\section{Background}\label{background}

\subsection{Case-Based Reasoning}\label{case-based-reasoning}

In case-based reasoning (CBR) \cite{aamodt1994, gilboa1995} a case is a tuple $C = (P, S, R)$ where $P= (p_1, p_2,\dots, p_n)$ denotes the problem made up of $n$ components, $S$ is the solution, and $R$ is the result of $S$. Let $\mathcal{C} = (C_1, C_2,\dots, C_N)$ denote a repository of $N$ solved cases. Given a case $C_i$ for $i \in [1,N]$ denote by $P_i$, $S_i$ and $R_i$ the problem, solution and result of $C_i$ respectively; furthermore, denote by $p_{i,j}$ the problem component $j \in [1,n]$ for case $i$. Let $\tilde{C} = (\tilde{P}, \phi, \phi)$ be a new case with $\tilde{P} = (\tilde{p}_1, \tilde{p}_2,\dots, \tilde{p}_n)$, where $\phi$ denotes that we do not yet have a solution or result for the new case. Then the CBR pipeline operates in four phases:

\textbf{Retrieve}: the system searches its case repository for cases that are similar to the new case. Formally, let $\rho(\tilde{P}, P_i) $ denote a similarity function that measures how similar the problem of cases $\tilde{C}$ and case $C_i$ are. Then the $k>0$ most similar cases, denoted $\mathcal{C}_k$, are given by:

\begin{equation}\label{eq:1}
\mathcal{C}_k =\operatorname*{arg\,top_k}_{C_i \in \mathcal{C}} \rho(\tilde{P}, P_i).
\end{equation}

\textbf{Reuse}: the retrieved cases are reused to find a solution to the new case. Formally, let $\textrm{Reuse}$ be a function that reuses the solutions of similar cases $\mathcal{C}_k$ to find a solution $\tilde{S}$ to $\tilde{P}$. Then

\begin{equation}\label{eq:2}
\tilde{S} = \textrm{Reuse}(\tilde{P}, \mathcal{C}_k).
\end{equation}

\textbf{Revise}: the result of the proposed solution is obtained and, if found to be incorrect, revised. Formally, let $R^{*}$ denote the correct result of $\tilde{C}$ and $\textrm{Eval}$ be a function that evaluates a given solution to produce a result. Then, if $\textrm{Eval}(\tilde{S})= \tilde{R} \neq R^{*}$, revise $\tilde{S}$ using a revision function $ \textrm{Revise}$:

\begin{equation}\label{eq:3}
\tilde{S} \leftarrow \textrm{Revise}(\tilde{S}),
\end{equation}

such that $\tilde{R}=R^{*}$.

\textbf{Retain}: once the new case is solved, add it to the repository of solved cases. Formally, if $\tilde{R}=R^{*}$ then add $\tilde{C}=(\tilde{P},\tilde{S},\tilde{R})$ to $\mathcal{C}$:

\begin{equation}\label{eq:4}
\mathcal{C} \leftarrow \mathcal{C} \cup \{ \tilde{C} \}.
\end{equation}

\subsection{Retrieval-Augmented Generation}

Let $x$ be a sequence of input tokens, and let $q = \textrm{Query}(x)$ represent a function that generates a query $q$ from $x$. Let $\mathcal{L}$ denote a Large Language Model (LLM), such that $y = \mathcal{L}(q)$ corresponds to the LLM's response given the query $q$.

Retrieval-Augmented Generation (RAG) \cite{lewis2020} is a technique designed to enhance the quality of the LLM's response by incorporating additional context from an external document memory. Specifically, given a collection of $M$ documents $\mathcal{D} = \{d_1, d_2, \dots, d_M\}$, the method retrieves the top $k>0$ most similar documents to the query, denoted $\mathcal{D}_k$, using:

\begin{equation}\label{eq:5}
\mathcal{D}_k =\operatorname*{arg\,top_k}_{d \in \mathcal{D}} \textrm{CosineSim}(\mathfrak{E}(q), \mathfrak{E}(d)),
\end{equation}

where $\mathfrak{E}$ is a function that maps text to an embedding space, and $\textrm{CosineSim}(a, b) = \frac{a \cdot b}{\|a\|\|b\|}$ is the cosine similarity between vectors $a$ and $b$. 

Once the relevant documents are retrieved, an augmented query can be constructed: $q_{RAG} = \textrm{Query}(x, \mathcal{D}_k)$. This query incorporates both the input tokens $x$ as well as $\mathcal{D}_k$ as additional context. The LLM then generates a response based on this augmented query:

\begin{equation}\label{eq:6}
y_{RAG} = \mathcal{L}(q_{RAG}).
\end{equation}

\section{Methodology}\label{methodology}

Research has demonstrated that RAG can be used to handle the first two phases of the CBR pipeline \cite{wiratunga2024, yang2024, guo2024, sourati2023}. In particular, if we consider cases that are text-only, then RAG retrieval, given by Equation \ref{eq:5}, can cater for the CBR Retrieve phase, given by Equation \ref{eq:1}, by converting case problems to embeddings. Meanwhile, RAG generation, given by Equation \ref{eq:6}, can cater for the CBR Reuse phase, given by Equation \ref{eq:2}, by passing an augmented query to an LLM that generates a solution to a new case and providing the retrieved solved cases as additional context.

Unfortunately, the Revise and Retain phases cannot be handled generally, as automatically inferring whether an LLM has generated an incorrect solution and then revising the solution accordingly is not necessarily straightforward.\footnote{Some applications are more amenable to this. For example, DS-agent is a CBR application that solves Machine Learning tasks \cite{guo2024}. This application can implement a Revise phase by evaluating how well a proposed solution performs on a test set, and then retaining cases with solutions that improve test set performance.} For the purposes of this paper, we focus on the Retrieve and Reuse phases in presenting our general framework, leaving the latter two phases as application specific.

To formalize MCBR-RAG, we assume a multimodal CBR setting where the components $p_j$ of a case problem $P= (p_1, p_2,\dots, p_n)$ each has arbitrary modality, while the solution. $S$, of a case is text-based. Let $\{\frak{T}_j\}_{j=1}^{n}$ be a collection of $n$ text generation functions, such that $\mathfrak{T}_j(p_j) = t_j$ returns a text-based representation of $p_j$. Let $\{\frak{L}_j\}_{j=1}^{n}$ be a collection of $n$ latent representation functions, such that $\mathfrak{L}_j(p_j) = l_j$ returns a latent representation of $p_j$.

\textbf{Retrieve}: to process this phase under our proposed setting, we convert the problem components of a case to their latent representations and compute a weighted average over the cosine similarities of the individual components.

\begin{equation}\label{eq:7}
\mathcal{C}_k = \operatorname*{arg\,top_k}_{C_i \in \mathcal{C}} \sum_{j=1}^{n} w_j \textrm{CosineSim}(\mathfrak{L}_j(\tilde{p}_j) ,\mathfrak{L}_j(p_{i,j})),
\end{equation}

where $w_j > 0$ and $\sum_{j=1}^{n}{w_j} = 1$. Using a weighted average allows control over the importance of the different problem components.

\textbf{Reuse}: to process this phase under our proposed setting, we convert the problem components of a case to their text-based representations and pass these as additional context to an LLM. 

Let $\mathfrak{T}(P) = (\mathfrak{T}_1(p_1), \mathfrak{T}_2(p_2),\dots,\mathfrak{T}_n(p_n))$ be a text-based representation of a case problem. Let $\mathfrak{T}(C) = (\mathfrak{T}(P), S, R)$ be a case whose problem has been converted to text. Then we can build an augmented query $q_{RAG} = \textrm{Query}(\mathfrak{T}(\tilde{P}), \mathfrak{T}(\mathcal{C}_k))$, where $\mathfrak{T}(\mathcal{C}_k) = \{\mathfrak{T}(C)\}_{C\in\mathcal{C}_k}$, and obtain a solution to a new case 

\begin{equation}\label{eq:8}
\tilde{S} = \mathcal{L}(q_{RAG}).
\end{equation}

\section{Applications}\label{applications}

\subsection{Math-24}\label{applications-math-24}

\begin{figure}[ht]
\centering
\includegraphics[height=1.2in]{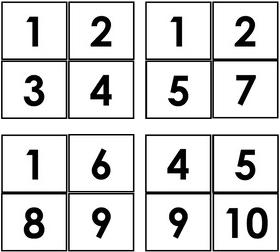}
\caption{Four Math-24 puzzle cards. The bottom right card is $(4,5,9,10)$ and has solutions $(10-4) \times (9-5)$ and $(10+5-9)\times 4$. }
\label{fig:math-24-examples}
\end{figure}

We first introduce a simplified Math-24 application that serves to illustrate the fundamental aspects of MCBR-RAG. A Math-24 puzzle is comprised of a card that contains $4$ numbers $(a ,b, c, d)$ with $1 \leq a \leq b \leq  c \leq  d \leq 13$. The goal is to use these four numbers to make $24$, where all four numbers must be used exactly once and one may use any of the addition, subtraction, multiplication and division operators. Figure \ref{fig:math-24-examples} shows four examples of Math-24 puzzles.

In total, there are $1362$ puzzles with at least one solution \cite{4nums_facts}. However, for the purposes of this application, we restrict our attention to the subset of puzzles that have a solution in the form $(x_1 \odot x_2) \times (x_3 \odot x_4) =24$ where $\odot \in \{+, -, \times, \div\}$. This is a common strategy to solve 
Math-24 puzzles and involves figuring out how to obtain $\{1, 24\}$, $\{2, 12\}$, $\{3, 8\}$, or $\{4, 6\}$ from pairs of puzzle numbers, then multiplying them together to get 24. This leaves us with a total of $466$ puzzles.

In transforming Math-24 into a CBR application, a case is defined as a tuple $(P, S, R)$. Here, $P$ represents an image of a Math-24 puzzle card; $S$ is a set containing the solutions to $P$,\footnote{In Section \ref{case-based-reasoning} we formally presented CBR as having a single solution per case. However, extending to a multi-solution setting is straightforward.}; and $R$ is a set of the results for these solutions, with each result being $24$ if the solution is correct. To handle this application within the MCBR-RAG framework, we require functions for both text generation and latent representation.

\begin{figure}[ht]
	\centering
	\begin{tikzpicture}
	
	\node(input) at (-1,0) [anchor=center, inner sep=0]{\includegraphics[width=1.6cm]{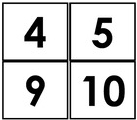}};
	\node at (-1,-1.25) {\scriptsize $90\times90$ Image};
	
	\node[draw, minimum width=3cm, minimum height=1.5cm, fill=blue!20] (convpool) at (2, 0) {};
	\node at (2,-1.25) {\scriptsize 3 x Conv + MaxPool};
		
	\node[draw, minimum width=0.5cm, minimum height=1.5cm, fill=blue!20] (flatten) at (4.5, 0) {};
	\node at (4.5,-1.25) {\scriptsize Flatten};
	
	\node[draw, minimum width=0.5cm, minimum height=1.5cm, fill=blue!20] (dense) at (5.5, 0) {};
	\node at (5.5,-1.25) {\scriptsize Dense};
	
	\node[draw, circle, fill=orange!50] (softmax1) at (8, 1) {};
	\node[draw, circle, fill=orange!50] (softmax2) at (8, 0.33) {};
	\node[draw, circle, fill=orange!50] (softmax3) at (8, -0.33) {};
	\node[draw, circle, fill=orange!50] (softmax4) at (8, -1) {};
	\node at (8, -1.5) {\scriptsize 4 x Softmax Outputs};
	
	\node (num1) at (9, 1)  {\scriptsize 4};
	\node (num2) at (9, 0.33) {\scriptsize 5};
	\node (num3) at (9, -0.33) {\scriptsize 9};
	\node (num4) at (9, -1) {\scriptsize 10};
	
	\draw[->] (input.east) -- (convpool);
	\draw[->] (convpool) -- (flatten);
	\draw[->] (flatten) -- (dense);
	\draw[->] (dense) -- (softmax1);
	\draw[->] (dense) -- (softmax2);
	\draw[->] (dense) -- (softmax3);
	\draw[->] (dense) -- (softmax4);
	\draw[->] (softmax1) -- (num1);
	\draw[->] (softmax2) -- (num2);
	\draw[->] (softmax3) -- (num3);
	\draw[->] (softmax4) -- (num4);
	
	\end{tikzpicture}

\caption{CNN for learning text generation in Math-24. Once the model is trained, the predictions can be used to generate a text-based representation of a Math-24 card image i.e. `4 5 9 10' for the image in this figure.}
\label{fig:math-24-cnn}
\end{figure}
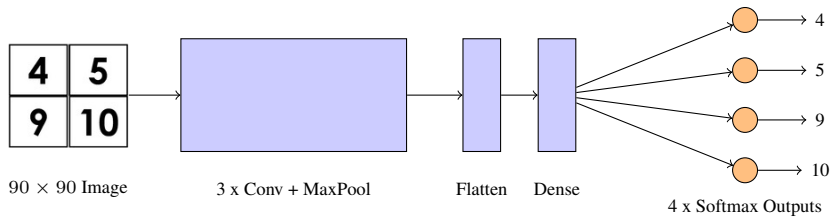

\textbf{Text generation}: We generate $40,000$ card images, each with $4$ random numbers between $1$ and $13$. We then train a convolutional neural network (CNN) that takes as input an image of a card and has four softmax outputs. Each softmax output has $13$ units that predicts a respective number on the card. Due the simple nature of this task, our model achieves $100\%$ accuracy on an independent test set of $10,000$ cards. Once the CNN is trained we can pass a Math-24 card image to the model and, from the resulting predictions, generate a text-based representation for the puzzle numbers on the card. See Figure \ref{fig:math-24-cnn} for an illustration of the CNN.

\textbf{Latent representation}: to learn a useful latent representation for this application, we train a fully-connected feedforward neural network (FFNN) treating the problem as a multi-label classification task. The input to the network is the text-based representation of the puzzle, which we predict using our text generator\footnote{In Section \ref{methodology} our notation suggested that the latent functions take in as input the modal problem components, but in many applications it is beneficial to work with the text-based representations from the text generation functions.}. Given our small dataset of only $466$ cases, we include additional aggregate features to aid classification. In particular, we count the number of pairs that can make $1$, $2$, $3$, $4$, $6$, $8$, $12$, or $24$ using any of the allowed operations. We do this both globally and per puzzle number. For example, consider the puzzle $(4,5,9,10)$. If we compute all pair combinations that can make $1$, we find that $(5-4)=1$ and $10-9=1$, so the global count of all pair combinations that can make $1$ is equal to $2$. We repeat this over all our candidates giving $8$ features. Next, suppose we take the first puzzle number, $4$, and compute which other puzzle numbers can be used with it to make $1$. Then we find that $(5-4)=1$, so the count of all pair combinations that can make $1$ with the first puzzle number is equal to $1$. We repeat this over all four puzzle numbers and all candidates giving $32$ features. In total, we have $40$ input features.

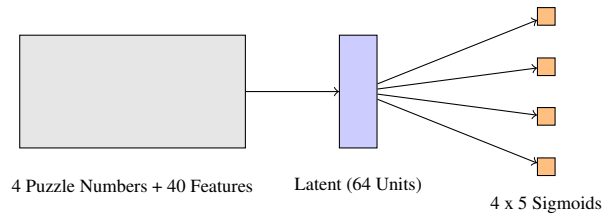
\begin{figure}[ht]
	\centering
	\begin{tikzpicture}
	
	\node[draw, minimum width=3cm, minimum height=1.5cm, fill=gray!20] (input) at (-1,0) {};
	\node at (-1,-1.25) {\scriptsize 4 Puzzle Numbers + 40 Features};
	
	\node[draw, minimum width=0.5cm, minimum height=1.5cm, fill=blue!20] (latent) at (2, 0) {};
	\node at (2,-1.25) {\scriptsize Latent (64 Units)};
	
	\node[draw, fill=orange!50] (sigmoid1) at (4.5, 1) {};
	\node[draw, fill=orange!50] (sigmoid2) at (4.5, 0.33) {};
	\node[draw, fill=orange!50] (sigmoid3) at (4.5, -0.33) {};
	\node[draw, fill=orange!50] (sigmoid4) at (4.5, -1) {};
	\node at (4.5, -1.5) {\scriptsize 4 x 5 Sigmoids};
	
	\draw[->] (input) -- (latent); 
	\draw[->] (latent) -- (sigmoid1); 
	\draw[->] (latent) -- (sigmoid2);
	\draw[->] (latent) -- (sigmoid3);
	\draw[->] (latent) -- (sigmoid4);
	\end{tikzpicture}

\caption{FFNN for learning latent representations in Math-24.}
\label{fig:math-24-ffnn}
\end{figure}

The output to the network consists of twenty sigmoid functions. Each batch of five sigmoids encodes 1) a solution category; and 2) the decomposition of the given solution category given by the location of the first two puzzle numbers used to make the larger number in the given solution category, in ascending order. For example, a solution $(10-4) \times (9-5)$ implies a solution category $\{4, 6\}$, and the first sigmoid in the batch of five would activate to encode this. Then, to make the larger number, $6$, we use $4$ and $10$ which are the first and fourth numbers of the puzzle. Thus, we can encode this solution as $[1,1,0,0,1]$. Since we have four solution categories, we need twenty sigmoid outputs in total. See Figure \ref{fig:math-24-ffnn} for an illustration of the FFNN.  

As a result of this architecture, the latent representation produced by this model can be used to find puzzles with both similar solution categories as well as similar decompositions, ensuring that similar retrieved cases are useful in solving new cases.  

\subsection{Backgammon}\label{applications-backgammon}

In this section, we introduce a more complex Backgammon application. Backgammon is a two-player game where players (O and X) move their pieces (called checkers) around triangles (called points) on a board. Given the current checker configuration (called a position), the player whose turn it is rolls a pair of dice, selects a move from the resulting legal set of moves and reconfigures the checkers accordingly. After playing their move, a new position is reached and the turn switches to the opposing player.

\begin{figure}[ht]
\centering
\includegraphics[height=2in]{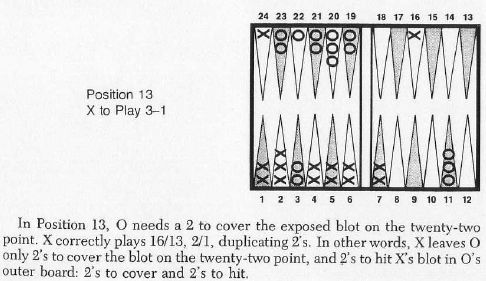}
\caption{An example of a Backgammon lesson \cite{magriel1997}. The board image represents the current position, and the image caption indicates that player X rolls a 3-1. The expert's analysis below the image discusses the move 16/13, 2/1.}
\label{fig:backgammon-lesson}
\end{figure}

Numerous books have been written on Backgammon by expert players. Such books often structure their lessons by showing an image of a board position and a roll for a given player, then provide an analysis of the pros and cons of a subset of legal moves. Figure \ref{fig:backgammon-lesson} presents a typical example \footnote{All Backgammon board images in this paper are taken from \cite{magriel1997}.}. 

Considering this typical structure, we can convert Backgammon lessons into a CBR applications by defining a case $C = (P, S, R)$. Here, $P=(z, p, r, m)$, where $z$ the image of the current position $p$ is the player on roll, $r$ is the rolled dice and $m$ is the move; $S$ represents the expert analysis of move $m$ given $z$, $p$ and $r$; and $R$ is a binary variable indicating whether the analysis is accurate. To build a repository of cases, we have collected the necessary data from three Backgammon books \cite{magriel1997, robertie2022, trice2004}, giving us a total of $1015$ cases. As with the previous application, we need functions for text generation and latent representation. 

\begin{figure}[ht]
\centering
\includegraphics[height=1.5in]{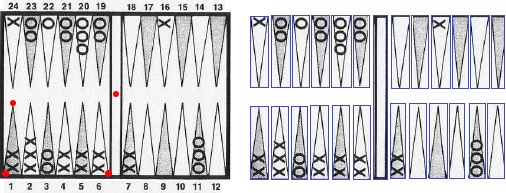}
\caption{Given the $4$ landmark coordinates, as shown by the red dots on the left image, we can slice up the image to obtain the $24$ points, as well as the centrally located bar point, as illustrated on the right. For example, given the bottom two coordinates, we compute the distance, $d$, between the start of the 1-point and the end of the 6-point along the $x$-axis. As there are $6$ points per quadrant in Backgammon, we can infer that the width of each point is $\frac{d}{6}$. We can then use this to find the start location of the 6-point along the $x$-axis. The other locations can also be inferred using similar logic.}
\label{fig:backgammon-landmarks}
\end{figure}

\textbf{Text generation}: to convert a Backgammon board image into a text-based representation, we break the problem up into two parts. We first note that, due to the symmetry in a Backgammon board, one only needs $4$ landmark coordinates to slice up the board into its constituent parts. See Figure \ref{fig:backgammon-landmarks} for an illustration.

\begin{figure}[ht]
\centering
\begin{tikzpicture}
\node(input) at (-1,0) [anchor=center, inner sep=0]{\includegraphics[width=1.8cm]{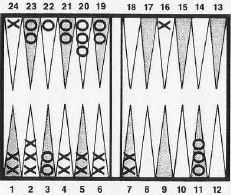}};
\node at (-1,-1.25) {\scriptsize $160\times160$ Image};

\node[draw, minimum width=3cm, minimum height=1.5cm, fill=blue!20] (convpool) at (2, 0) {};
\node at (2,-1.25) {\scriptsize 3 x Conv };

\node[draw, minimum width=0.5cm, minimum height=1.5cm, fill=blue!20] (flatten) at (4.5, 0) {};
\node at (4.5,-1.25) {\scriptsize Flatten};

\node[draw, minimum width=0.5cm, minimum height=1.5cm, fill=blue!20] (dense) at (6, 0) {};
\node at (6,-1.25) {\scriptsize 3 x Dense};

\node[draw, fill=orange!50] (linear1) at (9, 1) {};
\node[draw, fill=orange!50] (linear2) at (9, 0.33) {};
\node[draw, fill=orange!50] (linear3) at (9, -0.33) {};
\node[draw, fill=orange!50] (linear4) at (9, -1) {};
\node at (9, -1.5) {\scriptsize 8 x Linear};

\node (num1) at (10, 1) {\scriptsize $(x_1,y_1)$};
\node (num2) at (10, 0.33) {\scriptsize $(x_2,y_2)$};
\node (num3) at (10, -0.33) {\scriptsize $(x_3,y_3)$};
\node (num4) at (10, -1) {\scriptsize $(x_4,y_4)$};

\draw[->] (input.east) -- (convpool);
\draw[->] (convpool) -- (flatten);
\draw[->] (flatten) -- (dense);
\draw[->] (dense) -- (linear1);
\draw[->] (dense) -- (linear2);
\draw[->] (dense) -- (linear3);
\draw[->] (dense) -- (linear4);
\draw[->] (linear1) -- (num1);
\draw[->] (linear2) -- (num2);
\draw[->] (linear3) -- (num3);
\draw[->] (linear4) -- (num4);
\end{tikzpicture}

\vspace{0.5cm} 

\hspace*{-1.1cm}
\begin{tikzpicture}
\node(input1) at (-1,0) [anchor=center, inner sep=0]{\includegraphics[height=1.05cm]{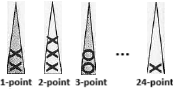}};
\node at (-1,-1.25) {\scriptsize Sequence of $24\times72$ Images};

\node[draw, minimum width=3.5cm, minimum height=1.5cm, fill=blue!20] (convpool) at (3, 0) {};
\node at (3,-1.25) {\scriptsize 3 x Conv + BatchNorm + MaxPool };

\node[draw, minimum width=0.5cm, minimum height=1.5cm, fill=blue!20] (flatten) at (5.5, 0) {};
\node at (5.5,-1.25) {\scriptsize Flatten};

\node[draw, minimum width=0.5cm, minimum height=1.5cm, fill=blue!20] (dense) at (7, 0) {};
\node at (7,-1.25) {\scriptsize 3 x Dense};

\node[draw, circle, fill=orange!50] (softmax1) at (9, 1) {};
\node[draw, circle, fill=orange!50] (softmax2) at (9, -1) {};
\node at (9, -1.5) {\scriptsize 2 x Softmax};

\node (num1) at (10, 1) {\scriptsize $0$};
\node (num2) at (10, -1) {\scriptsize $1$};

\draw[->] (input.east) -- (convpool);
\draw[->] (convpool) -- (flatten);
\draw[->] (flatten) -- (dense);
\draw[->] (dense) -- (softmax1);
\draw[->] (dense) -- (softmax2);
\draw[->] (softmax1) -- (num1);
\draw[->] (softmax2) -- (num2);
\end{tikzpicture}

\caption{CNNs for learning text generation in Backgammon. Once the models are trained, the predictions can be used to generate a text-based representation of a Backgammon board image i.e. `1X2 2X3 3O2 4X2 5X2 6X2 7X2 11O3 16X1 19O2 20O3 21O2 22O1 23O2 24X1' for the image in this figure.}
\label{fig:backgammon-cnn}
\end{figure}
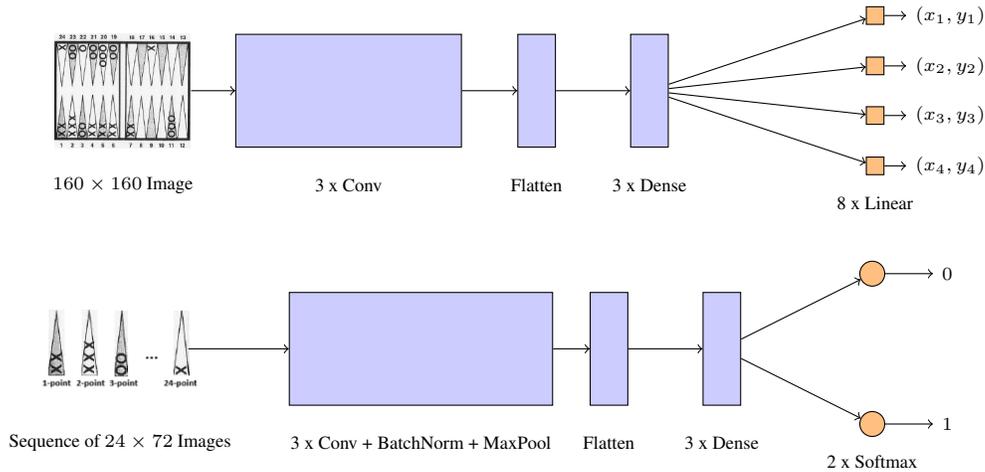

For our procedure, we simulate $100,000$ random Backgammon board images and train a CNN that takes in as input a board image and predicts the $4$ landmark coordinates. We then train a separate CNN that takes in as input a part image and predicts the number of checkers as well as the player who owns those checkers. At test time, given a new board image, we pass this image to the first CNN and get the landmark predictions. Next, we slice up the image into its constituent parts and predict each part with the second CNN. We then combine the part predictions to obtain a text-based representation of the board image. See Figure \ref{fig:backgammon-cnn} for an illustration.

This procedure has two sources of potential error: landmark prediction and part prediction. After converting a board image to text, we validate that the resulting position is legal and that the associated move can be legally applied. If one of these conditions fails, we discard the case. Our case retention rate with this procedure is approximately $90\%$.      

\begin{figure}[ht]
\centering
\begin{tikzpicture}
\node[draw, minimum width=3cm, minimum height=1.5cm, fill=gray!20] (input) at (-1,0) {};
\node at (-1,-1.25) {\scriptsize Encoded Position};

\node[draw, minimum width=0.5cm, minimum height=1.5cm, fill=blue!20] (latent) at (2, 0) {};
\node at (2,-1.25) {\scriptsize Latent (64 Units)};

\node[draw, minimum width=1cm, minimum height=0.5cm, fill=orange!50] (task1) at (5, 1) {};
\node at (5, 0.5) {\scriptsize Value};

\node[draw, minimum width=1cm, minimum height=0.5cm, fill=orange!50] (task2) at (5, -1) {};
\node at (5, -1.5) {\scriptsize Decoder};

\draw[->] (input) -- (latent); 
\draw[->] (latent) -- (task1); 
\draw[->] (latent) -- (task2);
\end{tikzpicture}

\caption{Multi-task FFNN for learning latent representations in Backgammon.}
\label{fig:backgammon-fnn}
\end{figure}
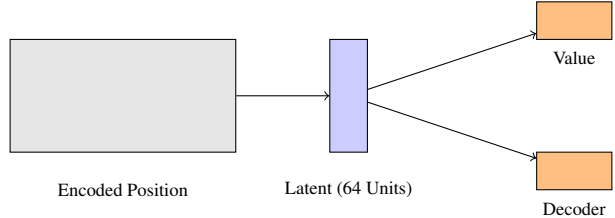

\textbf{Latent representation}: To learn a useful latent representation for this application, we train a multi-task FFNN. The input to the network is an encoded position, where we use an encoding similar to TD-Gammon \cite{tesauro1995}. For the output, we have two tasks: the first is the value of the position\footnote{In Backgammon, the value of a position is its win / loss probability. A public dataset with positions and their corresponding values has been released by GNU Backgammon \cite{gnubg_training}.} and the second is a decoder that predicts the encoded input. The value measures the relative positional strength between the two players, while the decoder ensures that the model optimizes for the precise checker configuration. In totality these two objectives encourage the model to consider positions as similar when they have both similar checker configuration and strategic value. See Figure \ref{fig:backgammon-fnn} for an illustration.

Once the FFNN is trained, then given a case with problem $P=(z, p, r, m)$ we can use our text generator to get a text-based representation, $t_z$, of $z$ and pass that to our latent model to get a latent representation, $l_z$. It is beneficial to also take into account the position after applying the move $m$. Let $g(z, m) = t_{z^\prime}$ be a function that returns $t_{z^\prime}$, the text-based representation of the position after applying move $m$ to position $z$. The corresponding latent representation is denoted $l_{z^\prime}$. Then we can measure case similarity by using $\frac{l_z + l_{z^\prime}}{2}$ as our combined latent representation.\footnote{This approach deviates from Equation \ref{eq:7}, which compares latent representations solely between individual problem components. However, extending this to functions over problem components is straightforward, and potentially useful across various applications. For example, in text-only applications it may be beneficial to concatenate the individual text components and compute a single embedding over the concatenated output, rather than average across individual component embeddings.}

\section{Experiments}\label{experiments}

In this section, we run experiments on our Math-24 and Backgammon applications introduced in Section \ref{applications} to evaluate retrieval and generation quality.\footnote{We provide a Python implementation for the Math-24 application, available at \href{https://github.com/OfirMarom/Math24}{https://github.com/OfirMarom/Math24}.} All our reported results average over $10$ independent runs where, in each run, we randomly hold back $30$ cases for testing. For retrieval quality we use the metrics Precision@k, Recall@k, F1-Score@k, NDCG@k (Normalized Discounted Cumulative Gain) and MRR@k (Mean Reciprocal Rank) where we average each metric over $k \in \{1, 2, 3, 4, 5\}$.

For Backgammon, in order to prevent data leakage when evaluating a test case, we exclude all solved cases whose moves originate from the same analysis as that test case. This ensures that our test cases are not influenced by their own analysis, which we assume to be unknown. 

\subsection{Math-24}

For retrieval quality, we consider two labeling schemes to determine if a retrieved case is relevant:

\begin{itemize} 
\item[a)] \textbf{Solution Category Only} (SCO): a retrieved case is classified relevant if it contains any solution category present in the test case. 
\item[b)] \textbf{Solution Category \& Decomposition} (SCD): a retrieved case is classified as relevant if it contains any solution category and decomposition present test case.
\end{itemize}

We also consider two similarity measures using different inputs for cosine similarity as described in Section \ref{applications-math-24}:

\begin{itemize} 
\item[a)] \textbf{Features}: this measure uses the $40$ input features of the retrieved case and test case.
\item[b)] \textbf{Latent}: this measure uses the latent representations of the retrieved case and test case.
\end{itemize}

\begin{table}[h!]
\centering
\begin{tabular}{llccccc}
\toprule
\textbf{Labeling} & \textbf{Similarity} & \textbf{Precision} & \textbf{Recall} & \textbf{F1-Score} & \textbf{NDCG} & \textbf{MRR} \\ 
\midrule

\multirow{2}{*}{SCO} 
               & Features & 71.9 & 1.5 & 2.9 & 72.6 & 81.5 \\ 
               & Latent   & \textbf{82.0} & \textbf{1.8} & 3.4 & 82.4 & 87.2 \\ 
\midrule
\multirow{2}{*}{SCD} 
               & Features & 45.4 & 4.2 & 7.1 & \textbf{46.5} & 58.1 \\ 
               & Latent   & \textbf{64.2} & \textbf{6.5} & 10.9 & \textbf{65.1} & 73.4 \\ 
\bottomrule
\end{tabular}
\caption{Retrieval quality metrics for Math-24 expressed as percentages (\%).}
\label{tab:retrieval-quality-math-24}
\end{table}

Table \ref{tab:retrieval-quality-math-24} shows that SCD labeling produces lower precision but higher recall compared to SCO labeling. For example, with latent similarity, SCD labeling achieves $64.2\%$ precision and $6.5\%$ recall compared to SCO labeling with $82.0\%$ precision and $1.8\%$ recall. This occurs because correctly classifying a case based on both solution category and decomposition is more challenging. However, SCD labeling also results in fewer total relevant cases per solution class, thus increasing recall.

Notably, across both labeling schemes, using latent similarity yields stronger performance on all metrics compared to feature similarity. For instance, with SCD labeling, latent similarity achieves an NDCG score of $65.1\%$, significantly higher than $46.5\%$ achieved by feature similarity. This demonstrates the effectiveness of leveraging latent representations, and justifies the additional steps required to learn a latent space for retrieval.

For measuring generation quality, we consider three query types:

\begin{itemize} 
\item[a)] \textbf{No Context} (NC): this query uses no context and asks an LLM to solve the puzzle directly.
\item[b)] \textbf{General Context} (GC): this query uses a general execution plan as context that applies to all puzzles in our test set.
\item[c)] \textbf{Top Context} (TC): this query uses the most relevant retrieved case as context to provide an LLM with a precise execution plan based on the solution category and decomposition of that case. 
\end{itemize}

An example of the three query types can be found in Appendix \ref{math-24-queries}. 

For this CBR application, it is straightforward to measure the result of an LLM solution as the final answer, if correct, must form an expression using all four puzzle numbers that evaluates to $24$.

\begin{table}[h!]
\centering
\begin{tabular}{llccc}
\toprule
\textbf{LLM} & \textbf{Query} & \textbf{Accuracy} & \textbf{Faithfulness} & \textbf{Negative Rejection} \\
\midrule
\multirow{3}{*}{Llama-3.1-405B-Instruct-Turbo} 
& NC & \textbf{34.0} & &  \\
& GC & \textbf{30.0} &  &  \\
& TC & 61.3 & 73.8 & 33.5 \\
\midrule
\multirow{3}{*}{Qwen2-72B-Instruct} 
& NC & \textbf{16.7} &  &  \\
& GC & \textbf{22.0} &  &  \\
& TC & \textbf{47.3} & 59.2 & 18.2 \\
\midrule
\multirow{3}{*}{Llama-3.1-70B-Instruct-Turbo} 
& NC & 17.0 &  &  \\
& GC & 25.3 &  &  \\
& TC & 45.3 & \textbf{56.2} & \textbf{18.8} \\
\midrule
\multirow{3}{*}{Llama-3-70b-chat-hf} 
& NC & 6.0 &  &  \\
& GC & 12.3 &  &  \\
& TC & 53.3 & 69.4 & 15.9 \\
\midrule
\multirow{3}{*}{Mixtral-8x7B-Instruct-v0.1} 
& NC & 7.3 &  &  \\
& GC & 4.0 &  &  \\
& TC & 14.7 & 20.5 & 2.1 \\
\midrule
\multirow{3}{*}{Llama-3.1-8B-Instruct-Turbo} 
& NC & \textbf{10.7} &  &  \\
& GC & 0.7 &  &  \\
& TC & 14.3 & 19.4 & 2.8 \\
\bottomrule
\end{tabular}
\caption{Generation quality metrics for Math-24 expressed as percentages (\%).}
\label{tab:generation-quality-math-24}
\end{table}

In Table \ref{tab:generation-quality-math-24} we report Accuracy, Faithfulness and Negative Rejection for a range of LLMs \cite{touvron2023, bai2023, jiang2024}.\footnote{Faithfulness measures how often a solution is correct given the context contains at least one correct solution, while Negative Rejection measures how often a solution is correct given the context contains no correct solutions.} Our results show that larger LLMs are generally better at solving Math-24 puzzles. For instance, \textit{Llama-3.1-405B-Instruct-Turbo} outperforms \textit{Llama-3.1-8B-Instruct-Turbo} by $23.3\%$ even when no context is provided. Across all LLMs, the TC query type consistently yields the highest accuracy. For example, \textit{Qwen2-72B-Instruct} demonstrates a $25.3\%$ accuracy improvement compared the GC query type and a $30.6\%$ improvement over the NC query type.

Interestingly, GC does not consistently outperform NC across all models. For instance, \textit{Llama-3.1-405B-Instruct-Turbo} has $4\%$ higher accuracy with NC compared to GC. This suggests that, while context can be beneficial, it may also lead LLMs to fixate on provided information, potentially overlooking alternative problem-solving strategies. Furthermore, Faithfulness is notably higher than Negative Rejection across all models. For example, \textit{Llama-3.1-70B-Instruct-Turbo} achieves $56.2\%$ Faithfulness while only managing $18.8\%$ Negative Rejection. This indicates that when accurate context is supplied, the LLM effectively follows the execution plan to find a solution. Conversely, when presented with incorrect context, the LLM struggles to bypass it and find a correct solution.

\subsection{Backgammon}

For retrieval quality, we label cases using the chapter names of the Backgammon book from which the cases originate. As each book uses different chapter names, we manually group similar chapters together. To ensure diversification, we only consider chapters that can be grouped with at least one other chapter. Using this approach, we are able to label approximately $75\%$ of our cases.\footnote{Given that each book follows a different curriculum, the grouping of similar chapters involves some level of overgeneralization, which may lead to cases being labeled together that do not align. Furthermore, it is possible that two cases in unrelated chapters are, in fact, similar and will be incorrectly counted as false negatives. Nevertheless, this approach allows us to label Backgammon lessons across different sources in a largely objective manner.} See Appendix \ref{backgammon-chapter-groupings} for our grouping categories. 
 
For similarity, we use the encoded board position as well as our latent representation as described in Section \ref{applications-backgammon}. For the latent representation we further consider different weights to the multi-task FFNN that control the importance of the value and decoder objectives when training the network. In particular we test weights $(0, 1)$, $(1, 0)$, $(1, 1)$, $(1, 0.1)$, $(1, 0.01)$, $(1, 0.001)$, $(1, 0.0001)$ for the value and decoder respectively. For example, the weights $(1, 0)$ will give a weight $1$ to the value objective and a weight of $0$ to the decoder objective, effectively disconnecting the decoder from the network.

\begin{table}[h!]
\centering
\begin{tabular}{llccccc}
\toprule
\textbf{Similarity} & \textbf{Precision} & \textbf{Recall} & \textbf{F1-Score} & \textbf{NDCG} & \textbf{MRR} \\ 
\midrule
Encoding & 35.3 & 2.5 & 4.4 & \textbf{36.6} & 46.6 \\ 
Latent(0, 1) & 36.6 & 2.6 & 4.6 & \textbf{38.0} & 47.9 \\ 
Latent(1, 0) & 44.4 & \textbf{3.5} & 6.1 & \textbf{45.5} & 53.7 \\ 
Latent(1, 1) & 40.9 & 3.0 & 5.3 & 42.2 & 51.8 \\ 
Latent(1, 0.1) & 40.9 & 3.1 & 5.5 & 42.3 & 51.8 \\ 
Latent(1, 0.01) & 44.4 & \textbf{3.4} & 6.0 & \textbf{45.6} & \textbf{54.2} \\ 
Latent(1, 0.001) & 43.1 & 3.4 & 6.0 & 44.0 & 52.7 \\ 
Latent(1, 0.0001) & 44.2 & 3.5 & 6.1 & 45.4 & 54.3 \\ 
\bottomrule
\end{tabular}
\caption{Retrieval quality metrics for Backgammon expressed as percentages (\%).}
\label{tab:retrieval-quality-backgammon}
\end{table}

Our results are presented in Table \ref{tab:retrieval-quality-backgammon}. We first note that Encoding and Latent(0, 1) exhibit the worst performance across all metrics. This is because they focus solely on checker configuration, ignoring value. However, these methods still demonstrate some retrieval capability with NDCG scores of $36.6\%$ and $38.0\%$ respectively. In contrast, Latent(1, 0), which incorporates only value, demonstrates stronger performance with an NDCG of $45.5\%$. This finding suggests that value is a more effective measure of position similarity compared to checker configuration.

Given that both value and checker configuration demonstrate retrieval capabilities, it makes sense to combine them. Experimenting with different weights, we find that Latent(1, 0.01) outperforms Latent(1, 0) with slightly higher NDCG and MMR scores of $45.6\%$ and $54.2\%$ respectively. While it has slightly lower recall of $3.4\%$ compared to Latent(1, 0), that has recall $3.5\%$, generation quality tends to benefit from systems that prioritize relevance of retrieved documents \cite{karpukhin2020}. Hence we consider Latent(1, 0.01) our best model for similarity and use it in our subsequent generation quality experiments. 

For generation quality, we consider two query types:

\begin{itemize} 
\item[a)] \textbf{No Context} (NC): this base query provides the LLM with a Backgammon position, encoded in JSON format, along with the player on roll, the dice rolled, and the move played. The query then asks the LLM to generate an analysis of the move.
\item[b)] \textbf{With Context} (WC): this query appends the base query with additional context from the top three most similar cases. This includes the same information as the base query for each retrieved case (position, player, dice, and move), along with the corresponding expert analysis.
\end{itemize}

One challenge in specialized applications, such as Backgammon, is that token-level evaluation metrics, such as BERTScore \cite{zhang2020}, may not be as effective. The problem with such metrics is that a response may be deemed similar merely due to its heavy use of jargon, even if the response itself lacks coherence. To overcome this, we employ sentence transformers that measure semantic similarity across sequences of words, providing a richer comprehension of textual meaning. We provide qualitative examples supporting this claim in Appendix \ref{backgammon-qualitative-examples}. Specifically, we compute cosine similarity between sentence embeddings with the following sentence transformer models: \textit{all-mpnet-base-v2} \cite{song2020}, \textit{all-roberta-large-v1} \cite{liu2019}, and \textit{all-MiniLM-L12-v1} \cite{wang2020}.

\begin{table}[h!]
\centering
\begin{tabular}{llccccc}
\toprule
\textbf{LLM} & \textbf{Query} & \textbf{BERTScore (F1)} & \textbf{MPNet} & \textbf{RoBERTa} & \textbf{MiniLM}  & \textbf{Avg} \\
\midrule
\multirow{2}{*}{Meta-Llama-3.1-405B-Instruct-Turbo} 
& NC & \textbf{83.9} & 60.4 & 63.9 & 58.3 & 60.9\\
& WC & \textbf{83.3} & 61.8 & 65.1 & 60.3 & 62.4\\
\midrule
\multirow{2}{*}{Qwen2-72B-Instruct} 
& NC & \textbf{83.9} & 58.0 & 62.3 & 55.5 & 58.6\\
& WC & \textbf{84.1} & 58.9 & 63.5 & 56.7 & 59.7 \\
\midrule
\multirow{2}{*}{Meta-Llama-3.1-70B-Instruct-Turbo} 
& NC & 84.0 & 58.3 & 61.5 & 55.2 & \textbf{58.3}\\
& WC & 84.1 & 63.2 & 66.5 & 61.7 & \textbf{63.8} \\
\midrule
\multirow{2}{*}{Llama-3-70b-chat-hf} 
& NC & 83.8 & 57.2 & 59.2 & 54.4 & 56.9\\
& WC & 84.1 & 62.7 & 65.8 & 61.3 & 63.3\\
\midrule
\multirow{2}{*}{Mixtral-8x7B-Instruct-v0.1} 
& NC & 83.0 & 58.3 & 62.4 & 56.9 & 59.2\\
& WC & 83.0 & 60.2 & 64.1 & 59.6 & 61.3 \\
\midrule
\multirow{2}{*}{Meta-Llama-3.1-8B-Instruct-Turbo} 
& NC & 83.3 & 57.0 & 60.7 & 55.1 & 57.6\\
& WC & 81.5 & 61.4 & 65.0 & 59.8 & 62.1\\
\bottomrule
\end{tabular}
\caption{Generation quality metrics for Backgammon expressed as percentages (\%). BERTScore uses \textit{roberta-large} for embeddings \cite{liu2019}. The Avg column is the average of MPNet, RoBERTa and MiniLM.}
\label{tab:generation-quality-backgammon}
\end{table}

Our results are presented in Table \ref{tab:generation-quality-backgammon}. We observe that BERTScore demonstrates inconsistent improvement for NC and WC query types. For example, \textit{Meta-Llama-3.1-405B-Instruct-Turbo} achieves a higher BERTScore with the NC query type, while \textit{Qwen2-72B-Instruct} performs better with WC query type. In contrast, \textit{all three} sentence transformer metrics consistently improve when context is provided compared to when it is absent. For example, \textit{Meta-Llama-3.1-70B-Instruct-Turbo}, the best-performing LLM, achieves an average score of $63.8\%$ with context, compared to $58.3\%$ without. These results indicate that generated answers with context are able to leverage this additional knowledge to produce higher-quality analysis of Backgammon positions.

\section{Final Remarks}\label{final-remarks}

This paper introduced MCBR-RAG, a general RAG framework designed for multimodal CBR applications. The framework addresses multimodal problem components through two core functions: a text generation function, which transforms problem components into text, and a latent representation function, which produces latent representations of these components. These functions facilitate the Retrieve and Reuse phases of the CBR pipeline through RAG retrieval and RAG generation, respectively.

We began by formalizing MCBR-RAG in an abstract setting to demonstrate its generalizability across various applications. This formalization provides a foundation for adapting the framework to a wide range of multimodal CBR scenarios. We then applied MCBR-RAG to two specific tasks: a simplified Math-24 application and a more complex Backgammon application, where we outlined practical methods for learning the necessary text generation and latent representation functions. Finally, our experiments demonstrated that, in both applications, MCBR-RAG produced higher generation quality across various LLMs compared to a baseline that did not incorporate contextual information in the LLM query.

\section*{Acknowledgments}

I would like to thank Professor Benjamin Rosman for his insightful suggestions and valuable discussions, particularly in framing this research as a broader methodology applicable to multiple domains.

\bibliographystyle{unsrt}
\bibliography{main.bbl}  

\appendix

\section{Math-24 Queries}\label{math-24-queries}

Consider the test problem $(1,3,7,12)$ with solution $(7-1)\times (12\div3)$. Suppose the most similar retrieved case is $(1,3,6,7)$ that has the following solutions: $(7-3) \times (6\times1)$ and $(7+1) \times (6-3)$. Then below are the three constructed queries.

\textbf{No Context:}
\begin{spverbatim}
START QUESTION
Solve the following Math-24 puzzle: 
1 3 7 12
END QUESTION
\end{spverbatim}

\textbf{General Context:}
\begin{spverbatim}
START QUESTION
Solve the following Math-24 puzzle:
1 3 7 12
END QUESTION

START CONTEXT
To help you answer the question, below is a tip that may help:
a) use a pair of numbers to make 24, 12, 8 or 6
b) use the remaining pair of numbers to make 1, 2, 3 or 4 respectively
c) then the product of step a) and step b) equals 24
END CONTEXT
\end{spverbatim}

\textbf{Top Context:} 

\begin{spverbatim}
START QUESTION
Solve the following Math-24 puzzle:
1 3 7 12
END QUESTION

START CONTEXT
To help you answer the question, below is a tip that may help:
a) use the pair (1, 7) to make 6. If this is impossible, try to make 6 using some other pair
b) then use the remaining pair to make 4
c) then 6 * 4 = 24
OR
a) use the pair (1, 12) to make 8. If this is impossible, try to make 8 using some other pair
b) then use the remaining pair to make 3
c) then 8 * 3 = 24
END CONTEXT
\end{spverbatim}

We note that in this example, only the first solution can be correctly applied to the test case. 

All queries use the same system prompt that encourages an LLM response to conform to a convention that we can easily parse to ascertain the correctness of its proposed solution.
 
\begin{spverbatim}
You are a student taking a test to solve a Math-24 puzzle. A Math-24 puzzle requires you to use 4 numbers to make 24. Each number must be used exactly once, and may use the + - * / operators. Once you have solved the puzzle, you must end your answer with "Final Answer: [LHS] = 24" where [LHS] is uses ALL 4 numbers EXACTLY once to get to 24. For example, if the puzzle is 1 2 9 13 then ending with "Final Answer: (13 + 9 + 2) * 1 = 24" will get you full marks. If you end your answer using any other convention you will get no marks, even if your final answer is correct. For example, if you end with "Final Answer: (13+9+2)=24; 24*1=24", this will get you no marks because your format is wrong, even though your answer is correct. If you end with "Final Answer: (13 + 9 + 2) = 24", this will also get you no marks because you have omitted the 1, even though your answer is correct. When giving your final answer, do not use any special formatting such as bold or italics, latex, etc. You must use only plain text.
\end{spverbatim}

\section{Backgammon Chapter Groupings}\label{backgammon-chapter-groupings}

\begin{longtable}{llc}
\toprule
\textbf{Book} & \textbf{Chapter} & \textbf{Group} \\
\midrule
\endfirsthead

\bottomrule
\caption{Chapter groupings for three Backgammon books.} \\
\endlastfoot

\multirow{18}{*}{\cite{magriel1997}} 
& Builders and Flexibility & Builders and Flexibility \\
& Duplication and Diversification & Duplication and Diversification \\
& When You Are Forced to Leave Shots & Risk and Safety \\
& Modern Opening Theory & Early Game Strategy \\
& Safe Play vs. Bold Play & Risk and Safety \\
& Slotting & Builders and Flexibility \\
& Action Play & Risk and Safety \\
& One Man Back & One Man Back \\
& Golden Point & Anchor Plays \\
& Splitting & Builders and Flexibility \\
& Doubling Theory & Doubling Theory \\
& No Possible Contact & Endgame No Contact \\
& Avoiding Contact & Endgame Against Contact \\
& After Contact & Containment \\
& Priming and Blocking & Priming Strategy \\
& Control of the Outside & Containment \\
& Timing & Timing \\
& Holding Game and Backgame & Backgames and Holding Games \\
\midrule
\multirow{21}{*}{\cite{robertie2022}} 
& The Opening & Early Game Strategy \\
& Flexibility & Builders and Flexibility \\
& The Blitz & Blitz Strategy \\
& One Man Back & One Man Back \\
& Holding Games & Backgames and Holding Games \\
& Priming Games & Priming Strategy \\
& Connectivity & Connectivity \\
& Hit or Not? & Hitting Decisions \\
& Breaking Anchor & Anchor Plays \\
& Action Doubles & Doubling Theory \\
& Late-Game Blitz & Blitz Strategy \\
& Too Good to Double & Doubling Theory \\
& Ace Point Games & Defensive Ace Point Games \\
& Back Games & Backgames and Holding Games \\
& The Containment Game & Containment \\
& Post-Ace-Point Games & Containment \\
& Playing for a Gammon & Gammon Strategy \\
& Saving the Gammon & Gammon Strategy \\
& Bearing off Against Contact & Endgame Against Contact \\
& The Race & Endgame No Contact \\
& The Bearoff & Endgame No Contact \\
\midrule
\multirow{15}{*}{\cite{trice2004}} 
& Primes And The Closeout & Priming Strategy \\
& Blitz & Blitz Strategy \\
& Anchors & Anchor Plays \\
& The Defensive Ace Point & Defensive Ace Point Games \\
& High Anchors And Holding Games & Backgames and Holding Games \\
& The Opening Roll & Early Game Strategy \\
& Blots, Shots, And Odds & Risk and Safety \\
& To Hit Or Not To Hit? & Hitting Decisions \\
& Staying Connected & Connectivity \\
& Bringing It Home & Endgame Against Contact \\
& Gaining Time & Timing \\
& Duplication & Duplication and Diversification \\
& Diversification & Duplication and Diversification \\
& Backgames & Backgames and Holding Games \\
& Primes & Priming Strategy \\
\end{longtable}

\section{Backgammon Qualitative Examples}\label{backgammon-qualitative-examples}

\subsection*{Example 1}

This example serves to illustrate that high similarity with sentence transformers correlates with high-quality generated Backgammon analysis. Consider the following actual analysis from \cite{magriel1997} followed by a generated answer with high sentence transformer score of $77.3\%$. 

We see that both analyses focus on the same underlying strategic concept of maximising coverage so that, if O tries to escape, X has a double shot.

\textbf{Actual:}
\begin{spverbatim}
Position 1 is a typical case which illustrates both these principles. The correct play is 17/14. By keeping your men spread out, you insure at least a double shot should O run with a 6. Although you may be tempted to button up to block 6-6 by playing 16/13, this would be a serious mistake. In positions where you are trying to catch one last man, it is much more important to be diversified to insure getting the maximum number of return shots, rather than to try to avoid being hit at all costs. By playing 16/13; you may only get a single direct shot at O if he runs with a 6 on his next roll.
\end{spverbatim}

\textbf{Generated:}
\begin{spverbatim}
In this position, X's move 17/14 is correct. The main goal is to maximize coverage and create the possibility of a double shot when O runs. By moving 17/14, X is insuring maximum coverage and putting pressure on O's potential rollouts. Moving 17/14 also takes into account O's potential rolls, such as 6-1, which would force O to come into direct range of X's men on the 14 and 16 points. This move is in line with the expert analysis from Example 1, where 18/14 is considered the correct play to ensure maximum coverage and create a double shot opportunity. In contrast, moving 17/13 would not achieve the same level of coverage and would leave X's men more exposed to O's potential rolls. Therefore, 17/14 is the correct play in this position.
\end{spverbatim}

\subsection*{Example 2}

This example serves to illustrate that high similarity with token-level metrics, like BERTScore, does not correlate with high-quality generated Backgammon analysis. Consider the actual analysis from \cite{robertie2022} followed by a generated answer with high BERTScore of $86.6\%$ and low sentence transformer score of $46.7\%$. 

We see that while the analyses contain similar Backgammon jargon, resulting in high BERTScore, the underlying explanations are very different. The actual analysis criticizes the move 8/4, 6/4, as it gives O a slim chance to win with a lucky roll next turn. Meanwhile, the generated answer misses this completely and praises the move for extending X's prime. The low sentence transformer score reflects this deeper semantic mismatch.

\textbf{Actual:}
\begin{spverbatim}
Weigh risks carefully. Black can play 8/4 6/4, or 13/9 11/9. The 4-point is a better point than the 9-point, without doubt. But it comes at a price - allowing White a 17-1 shot at Black's loose blot. Is it worth it? No. A 17-1 shot means giving your Opponent an extra 5.5% chance to win the game right on the spot. Making a better point picks up some winning chances, but not nearly that much. Play completely safe here by making the 9-point.
\end{spverbatim}

\textbf{Generated:}
\begin{spverbatim}
A fascinating position! With the roll of 4-2, X has made a solid move by playing 8/4 and 6/4. This decision aims to improve X's prime, creating a strong foundation for future rolls. By moving two checkers from the 8-point to the 4-point, X is building a flexible anchor that can be utilized to attack O's checkers or create a potential escape route. The secondary effect of this move is to slightly weaken O's anchor on the 12-point, as X now has more options to potentially hit an O checker on the 11-point. Additionally, X's 4-point anchor makes it more difficult for O to establish a strong prime in the future. Overall, this move is a well-thought-out strategic decision, as it improves X's board control, flexibility, and attacking potential while limiting O's options.
\end{spverbatim}

\end{document}